\RequirePackage{amsmath}
\documentclass[runningheads]{llncs}
\usepackage[T1]{fontenc}
\usepackage{graphicx}
\usepackage{hyperref}
\usepackage{color}

\usepackage{booktabs}

\begin{document}
\title{Improving Stance Detection by Leveraging Measurement Knowledge from Social Sciences: A Case Study of Dutch Political Tweets and Traditional Gender Role Division}
\titlerunning{Improving Stance Detection by Leveraging Measurement Knowledge}
\author{Qixiang Fang\orcidID{0000-0003-2689-6653} \and
Anastasia Giachanou\orcidID{0000-0002-7601-8667} \and
Ayoub Bagheri\orcidID{0000-0001-6366-2173}}
\authorrunning{Q. Fang et al.}
\institute{Department of Methodology and Statistics, Utrecht University, Utrecht, Netherlands \\
\email{\{q.fang,a.giachanou,a.bagheri\}@uu.nl}}
\maketitle              % typeset the header of the contribution
\begin{abstract}
Stance detection concerns automatically determining the viewpoint (i.e., in favour of, against, or neutral) of a text's author towards a target. Stance detection has been applied to many research topics, among which the detection of stances behind political tweets is an important one. In this paper, we apply stance detection to a dataset of tweets from official party accounts in the Netherlands between 2017 and 2021, with a focus on stances towards traditional gender role division, a dividing issue between (some) Dutch political parties. To implement and improve stance detection of traditional gender role division, we propose to leverage an established survey instrument from social sciences, which has been validated for the purpose of measuring attitudes towards traditional gender role division. Based on our experiments, we show that using such a validated survey instrument helps to improve stance detection performance.

\end{abstract}

\section{Introduction}
\label{intro}
Stance detection is an important natural language processing (NLP) task. Stance detection aims to automatically determine the viewpoint of a text's author towards a target~\cite{mohammad2016semeval}. This viewpoint, also known as stance, can be \textit{in favour of}, \textit{against} or \textit{neutral}. Typical targets include various social and political topics (e.g., abortion, feminism), political figures (e.g., Trump, Clinton) and events (e.g., referendums)~\cite{lahoti2018joint,lai2020multilingual,Aldayel2021}. 
Stance detection has been extensively applied to various research areas, including political discourse analysis~\cite{gunhal2022stance,lai2020multilingual}. Political tweets, in particular, serve as a rich source of information for understanding political stances and attitudes, making stance detection particularly relevant in this domain.

Detecting stances behind political tweets is a multifaceted task that involves analysing the linguistic nuances and contextual factors embedded within tweets to determine the viewpoint expressed towards a specific issue or target. Traditionally, this process has been approached using supervised learning methods, where a classifier is trained on a dataset of labelled tweets, where each tweet is associated with a predefined stance category~\cite{li2021p,samih2021few,lai2020multilingual}. 
Unsupervised learning methods for stance detection also exist, which do not rely on labelled data, but instead, seek to identify latent patterns and structures within a dataset of unlabelled tweets~\cite{gambini2023tweets,darwish2020unsupervised,stefanov2020predicting}. They typically employ techniques such as topic modelling, sentiment analysis, and distributional semantics to extract latent features from the tweets. These features are then analysed to identify clusters or groupings of tweets that share similar linguistic characteristics and semantic associations. These clusters are then assigned stance labels based on their perceived sentiment or topic orientation.

Recently, stance detection is increasingly conceptualised as textual entailment recognition (TER)~\cite{Aldayel2021,mohammad2016semeval}. TER, a more generic natural language understanding task, concerns determining whether a hypothesis follows from a premise, where three outcomes are possible: ``entailment'', ``contradiction'' and ``neutral''~\cite{dagan2004probabilistic}. By reformulating stance detection as judging whether a text (i.e., a premise) entails a stance towards a target (i.e., a hypothesis), a stance detection problem becomes a TER one. This reformulation offers two potential benefits. 
First, TER-based models are not target-specific and may be applied to unseen targets. Second, TER datasets can be used for training stance detection models. Currently, labelled stance detection datasets are limited to a few pre-defined targets and languages~\cite{Aldayel2021}. In contrast, TER datasets are not target-specific and are available in more languages. Therefore, we can use TER datasets (in addition to stance detection ones) for model training when considering stance detection as TER. Research has shown that training on additional TER datasets improves downstream stance detection performance~\cite{schiller2021stance}. 

In this paper, we investigate the application of stance detection as TER to a dataset of tweets from political parties and politicians in the Netherlands between 2017 and 2021, focusing specifically on stances towards traditional gender role division. This issue holds significant political and social implications, often dividing various Dutch political parties. To further improve stance detection performance and capture nuanced stances, we propose utilising a validated survey instrument from social sciences specifically designed to measure attitudes towards traditional gender role division. Our experiments demonstrate that incorporating this instrument indeed leads to improved stance detection performance. To the best of our knowledge, our study is the first to incorporate survey instruments for stance detection.

\section{Background}
\subsection{Stance Detection as Textual Entailment
Recognition}
TER, as mentioned in Section \ref{intro}, concerns judging whether a hypothesis ($h$) can be inferred from a premise ($p$). If yes, we conclude that $p$ entails $h$ ($p \Rightarrow h$). Typically, three outcomes are possible: ``entailment'', ``contradiction'' and ``neutral''. Note that the relation between $p$ and $h$ is directional, where $p \Rightarrow h$ does not imply $h \Rightarrow p$. 

% In stance detection, the task is to determine the stance of a text towards a target, where the stance can be ``in favour of'', ``against'' or ``neutral''.
The connection between the two tasks, TER and stance detection, becomes clear when we reformulate stance detection as judging whether a text ($t$) entails a \textbf{\textit{s}}tance \textbf{\textit{t}}o a \textbf{\textit{t}}arget (i.e., $t \Rightarrow stt$).
For instance, say the text $t$ is ``I appreciate more women in government positions'', the target is ``traditional gender role division'', and the stance is ``in favour of''. Then, we can formulate the stance to the target $sst$ as ``I am in favour of traditional gender role division''. As such, the task becomes judging whether ``I appreciate more women in government positions'' \textit{entails}, \textit{contradicts} or \textit{is neutral} towards ``I am in favour of traditional gender role division''. The three possible TER outcomes ``entailment'', ``contradiction'' and ``neutral'' correspond to the three stance detection outcomes: ``in favour of'', ``against'' and ``neutral''. 

One benefit of considering stance detection as TER is that a model can be easily trained on data involving multiple targets and can incorporate information about the targets during training. This may allow the model to predict stances towards unseen targets, also known as zero-shot learning~\cite{yin2019benchmarking}.
Another benefit is that TER datasets can be used for model training. For languages where existing stance detection datasets are available, additional TER datasets may help to further improve model performance. When stance detection datasets do not exist for a language but TER datasets do, we can still train a model based on TER datasets to predict stances. 
It is, however, important to note that current TER datasets are quite different from stance detection datasets~\cite{bowman-etal-2015-large,williams2018broad}.
The former mainly consists of descriptions of concrete scenes (e.g., ``two dogs are fighting'') which assume little world knowledge (e.g., social and political topics), while the latter are typically user-generated texts (e.g., social media posts) and concern diverse targets that can be abstract and ambiguous. 
This data discrepancy may limit the benefit of modelling stance detection as TER.

\subsection{Stance Detection as Attitudinal Measurements in Social Sciences}
In social sciences, stance detection is treated as a measurement problem and typically relies on the use of survey instruments. 
For instance, to measure stances towards traditional gender role division, one can ask a participant to indicate on a Likert scale from 1 to 5 how much they agree with the following statement:
``\textit{Overall, family life suffers the consequences if the mother has a full-time job.}''~\cite{Stam2014DoVM}.
The more they agree with this statement, the more they are in favour of the target (i.e., traditional gender role division).

Survey instruments need to be both ``\textit{valid}'' and ``\textit{reliable}''.
``Valid'' means that they measure what they are supposed to, while ``reliable'' means that the measurements do not suffer from large random variations~\cite{Trochim2015ResearchMT}. 
% Only when the survey instruments have demonstrated sufficient evidence for validity and reliability, can they measure stances accurately. 
One common way to establish the validity of survey instruments is to ensure that they cover all the aspects of a given target. This is also known as ``content validity''~\cite{Trochim2015ResearchMT}. 
For instance, when measuring stances to traditional gender role division, it is important to cover relevant subdomains of this target like in career, childcare, housework and education~\cite{Walter2018TheAO}.  
Furthermore, higher reliability can be achieved by averaging the responses to multiple survey instruments concerning the same target.

Established survey instruments exist for many stance detection targets (e.g., feminism~\cite{frieze1998measuring}, abortion~\cite{hendriks2012scale}, climate change~\cite{dijkstra2012development} and politicians~\cite{martin2005review}). They can potentially benefit the modelling of stance detection as TER.
For instance, when using a model trained on TER datasets to detect stances towards traditional gender role division, the  hypothesis can be as simple as ``I am in favour of traditional gender role division''. This hypothesis, however, does not explicitly cover all the relevant subdomains of the target (i.e., in career, childcare, housework and education). Consequently, the model may fail to identify texts that entail these subdomain stances.
We may mitigate this issue by using established survey items (that have demonstrated content validity) to construct multiple hypotheses.
Averaging the model's predictions across these multiple hypotheses can also lead to more reliable, less random results. 
Therefore, constructing multiple hypotheses based on established survey instruments for a given target may improve prediction validity and reliability. 

\section{Experiments}
We investigate the approach of modelling stance detection as TER by applying it to a dataset of tweets from Dutch political parties and politicians in the Netherlands, with the goal to measure stances towards traditional gender role division. Specifically, we conduct two experiments in a zero-shot stance detection setting. In the first experiment (Section~\ref{experiment1}), we fine-tune Bertje~\cite{devries2019bertje}, a Dutch BERT model, using Dutch TER data. Furthermore, we conduct a second experiment (Section~\ref{experiment2}) to validate the results from the first experiment. Given the recent developments and promising applications of large generative language models in various NLP tasks, we prompt GPT-3.5 Turbo, a large language model, to perform zero-shot stance detection. 

\subsection{Datasets}
We use Twitter (now X)'s API to collect a dataset of tweets from the official Twitter accounts of 13 major Dutch political parties (see Table~\ref{tab:ranking}) and the accounts of 10 politicians per party. We collect up to 3,200 most recent tweets per account (API limit) and in total, 311,362 tweets. We lower-case the tweets, remove urls and special characters, leave out tweets shorter than five words, and retain only the ones posted between 2017 and 2021. The resulting dataset contains 247,902 tweets. To reduce the number of tweets unrelated to the target of interest, we further apply a simple filtering strategy using keywords related to the target,\footnote{The Dutch equivalent of ``woman'', ``man'', ``mother'', ``father'', ``boy'' and ``girl'': ``vrouw'', ``man'', ``moeder'', ``vader'', ``jongen'', ``meisje''.} leading to a much smaller final dataset of 2,601 tweets. 

Previous research suggests that a party's stance towards an issue aligns with the average opinion of its voters~\cite{ibenskas2022congruence}. This inspires us to use the 2017-2021 data from the LISS panel\footnote{https://www.lissdata.nl/about-panel} to provide ground truth labels on the party level. The LISS panel is a representative sample of Dutch households that participate in monthly surveys about various topics like health, work and political views. The LISS data contains responses to 11 established survey items about stances towards traditional gender role division (see Section~\ref{app:chap7:survey}), as well as the party for which the participant last voted. 
We group the participants by the party they voted for and average their responses according to each participant's respective sampling weight across the 11 survey items for a given year and across all years. 
These scores are used to rank the parties in terms of their year(s)-specific stances towards traditional gender role division (see Table~\ref{tab:ranking}). For tweet-level ground truth labels, we rely on manual annotation (see Section~\ref{results} for more information, and Appendix~\ref{app:chap7:examples} for annotated examples). 

\begin{table}[htbp]
\caption{Parties Ranked by LISS-based Scores on Stances towards Traditional Gender Role Division. The higher the scores, the more in favour of traditional gender role division. Note that these scores are averaged across the years 2017-2021.}
\label{tab:ranking}
\centering
\begin{tabular}{@{}c|c@{}}
\toprule
Party      &  Score \\ \midrule
SGP        &  2.86      \\
DENK       &  2.56      \\
PVV        &  2.47      \\
PLUS50     &  2.38      \\
CU         &  2.35        \\
CDA        &  2.28        \\
FVD        &  2.26        \\
SP         &  2.16        \\
VVD        &  2.14        \\
PVDD       &  2.03        \\
PVDA       &  1.98        \\
D66        &  1.96    \\
GROENLINKS &  1.83        \\\bottomrule
\end{tabular}
\end{table}

Furthermore, we use the SICK-NL dataset~\cite{wijnholds-etal-2021-sicknl}, which is the Dutch translation of the original English SICK dataset~\cite{marelli-etal-2014-sick}. 
It contains 9,840 pairs of sentences that describe concrete scenes (e.g., ``Two dogs are fighting''), including 5,595 neutral pairs, 1,424 contradiction pairs and 2,821 entailment pairs.
To the best of our knowledge, SICK-NL is the only Dutch TER dataset. Also, no labelled stance detection dataset exists for the Dutch language.

\subsection{Experiment 1: Zero-shot Stance Detection with BERTje}
\label{experiment1}
Due to the lack of labeled Dutch stance detection datasets, we choose to fine-tune a TER classifier on the available SICK-NL dataset, and subsequently apply the trained classifier to perform zero-shot prediction of stances towards traditional gender role division.
Specifically, we fine-tune BERTje, a pretrained Dutch BERT model~\cite{devries2019bertje}, on the SICK-NL dataset, using 9,345 pairs for training and the rest for validation. We use the same hyperparameters as \cite{wijnholds-etal-2021-sicknl}. See Appendix~\ref{app:chap7:reprodu} for more computational details.
Next, we apply the fine-tuned BERTje model to our dataset of Dutch political tweets, where the goal is to predict whether a tweet entails a hypothesis (i.e., a stance towards traditional gender role division). We use two types of hypotheses. The first is a simple statement (in Dutch): ``I am in favour of traditional gender role division''. The second consists of the 11 survey items about traditional gender role division from the LISS panel (also in Dutch). See Section~\ref{app:chap7:survey} for more information about the construction of hypotheses.

We evaluate the stance detection performance at both the individual tweet level and the party level. Due to the large number of tweets in the dataset and lack of resources for manual annotation of all the tweets, we focus on evaluating only the top $k$ tweets with the highest predicted entailment scores, where $k = {10, 50, 100}$. 
The more out of the $k$ tweets are, indeed, true entailments, the more evidence that the model has encoded information useful for (zero-shot) stance prediction.

For the tweet-level evaluation, we obtain the labels for the $k$ tweets by asking two Dutch-speaking PhD researchers to manually annotate the tweets. Where inconsistent annotations occur, the annotators discuss them and agree on one correct label. Then, we calculate the percentages of correctly classified entailments among the $k$ tweets (i.e., top-$k$ precision) across two conditions (i.e., whether a simple hypothesis or survey items are used). 
We contrast these results with a random baseline, which is based on a random sample of 100 tweets from all the tweets in the final dataset. The same two PhD researchers from before manually annotate these 100 tweets. The percentage of entailments is used as the baseline precision score. 

For the party-level evaluation, we compute the average entailment probabilities for each party and year combination. Then, we calculate the
Spearman's correlation ($\rho$) between the average entailment probabilities and the LISS-based ranks of parties (across year and party). Scores close to 1 indicate better agreement.

\begin{table*}[htbp]
\caption{Top-$k$ precision scores for entailments, as well as
top-$k$ Spearman's $\rho$ between the estimated entailment probabilities and the LISS-based scores averaged across party and year. Top-$k$ refers to the tweets with the top 10/50/100 predicted entailment probabilities. The random baseline is based on a random sample of 100 tweets. The two non-baseline conditions concern whether a simple hypothesis (i.e., Without Survey) or survey items (i.e., With Survey) are used.}
\label{tab:results}
\centering
\begin{tabular}{@{}ccccccc@{}}
\toprule
& Random Baseline & Without Survey & With Survey 
\\\midrule   
$\text{Precision}_{10}$ & 0.11 & 0.10 & \textbf{0.70} \\
$\text{Precision}_{50}$ & 0.11 & 0.20 & \textbf{0.52} \\
$\text{Precision}_{100}$ & 0.11 & 0.22& \textbf{0.42} \\\midrule
Spearman's $\rho_{10}$   & -   & -0.04  & -0.07  \\
Spearman's $\rho_{50}$   & -   & 0.06   & -0.07  \\
Spearman's $\rho_{100}$  & -   & 0.15   & -0.08  \\
Spearman's $\rho_{all}$  & -   & 0.13   & -0.09  \\\bottomrule
\end{tabular}
\end{table*}

\paragraph{Results}
\label{results}
Table~\ref{tab:results} summarises the results on top-$k$ precision and Spearman's $\rho$. For the rows pertaining to top-$k$ precision, the scores are the percentages of tweets correctly classified as entailments. We can make two observations.
First, on the tweet level, the best precision scores consistently occur when survey items are used, indicating that the use of survey items to construct multiple hypotheses can be beneficial for stance detection. 
Second, however, the Spearman's $\rho$ scores paint a different picture: all the scores are very far from 1, indicating poor performance on the party level. 
These findings suggest that our Bertje model encodes useful information for zero-shot prediction of an unseen stance target, but only when survey items are used.

\subsection{Experiment 2: Zero-shot Stance Detection with GPT-3.5 Turbo}
\label{experiment2}
To further validate our findings from Experiment 1, we conduct a second experiment where we use GPT-3.5 Turbo, a large language model, to perform zero-shot stance detection of traditional gender role division.
We use a balanced dataset of 200 political tweets that were labelled during the annotation procedure of Experiment 1. We use the OpenAI API to prompt GPT-3.5 Turbo, with the following prompt template, where \{premise\} and \{hypothesis\} refer to two separate string variables:

\begin{quote}
``````

Given the following premise and hypothesis (both in Dutch), determine whether the premise entails, contradicts or is neutral about the hypothesis.

Premise: \{premise\}

Hypothesis: \{hypothesis\}

Note that the premise is an official tweet by a political party in the Netherlands, 
while the hypothesis is related to a positive stance towards traditional gender role division.
If the relationship between the premise and the hypothesis is entailment, this means that the political party behind the 
tweet/premise likely agrees with the hypothesis.

Give your answer response as one of the three outcomes: ``entails'', ``contradicts'' or ``neutral''.

''''''
\end{quote}

We use an English template because in this case, GPT-3.5 Turbo gives responses that are consistent with the requirement of the prompt (i.e., returning only ``entails'', ``contradicts'' or ``neutral''). In contrast, using a Dutch template would throw off the model, leading it to return sentences (e.g., ``The premise entails the hypothesis''; ``There is no relationship between the premise and the hypothesis'') and sometimes, responses in other languages like English.
To improve reproducibility, we set the temperature parameter to be 0 and the random seed to 1.

\paragraph{Results}
Using only a simple hypothesis, we obtain an accuracy score of 0.580; using all the survey items, however, we obtain a higher accuracy score of 0.725. This further suggests evidence for the efficacy of using survey items for stance detection. 

\section{Conclusion \& Discussion}
In this paper, we apply stance detection to a dataset of tweets from official party accounts in the Netherlands between 2017 and 2021, with a
focus on detecting stances towards traditional gender role division. We also investigate whether using established survey instruments can be helpful for stance detection.

In Experiment 1, we are faced with the challenge of lacking labelled stance detection datasets in Dutch. We thus opt to reformulate our stance detection task as TER, where we train a BERTje model on a Dutch TER dataset and subsequently apply it to zero-shot stance detection. We find using survey items consistently outperforms using a simple hypothesis for tweet-level stance detection. However, our model fails to achieve good stance detection on the party level.
In Experiment 2, we further test the benefit of using survey instruments by repeating the tweet-level analysis from Experiment 1 using prompts with OpenAI's GPT-3.5 Turbo API. We again find using survey questions beneficial for stance detection of traditional gender role division.

Our study has limitations. For instance, we focus on a single stance detection target and only a subset of the political tweet dataset. Including more stance detection targets and more datasets is necessary to further support the generalisability of our findings. Nevertheless, we hope that our study will inspire stance detection researchers to explore the use of survey instruments for stance detection research further. 
Furthermore, our second experiment, which utilised GPT-3.5, is now somewhat outdated given the recent emergence of more capable large language models like GPT-4 and GPT-4o. We also do not incorporate with other prompting engineering techniques such as chain-of-thought and few-short prompting, which have the potential to improve the model's task performance.

For future research, we suggest, among others, converting existing stance detection datasets into TER-compatible formats for model training, developing more general TER datasets (beyond concrete scenes), using multilingual models (where datasets in different languages may be combined), and comparing different strategies for constructing TER hypotheses from survey items. 

\begin{credits}
\subsubsection{\ackname}
We thank Dong Nguyen for her constructive and helpful feedback. 
This work was partially supported by the Dutch Research Council (NWO): grant number: VI.Vidi.195.152 to D. L. Oberski.

\subsubsection{\discintname}
The authors have no competing interests to declare that are
relevant to the content of this article. 
\end{credits}

\bibliographystyle{splncs04}
\bibliography{bibliography.bib}

\appendix
\section{Additional Reproducibility Information}
\label{app:chap7:reprodu}
\paragraph{Code and Data Availability} 
All the code and (anonymised) datasets will be made public on GitHub upon paper publication.

\paragraph{Computing Infrastructure}
All analyses were done on one of the researchers' personal laptop, which runs on an 11th Gen Intel(R) Core(TM) i7-11800H processor with 32 GB of RAM. 

Python 3.9, PyTorch 1.10.2, Huggingface Transformers 4.16.2 and CUDA 11.3 were used for finetuning BERTje and the downstream stance detection. R 4.1.2 and Tidyverse 1.3.1 were used for data cleaning and wrangling. 

\paragraph{Runtime}
The runtime for finetuning the BERTje model is about four GPU hours. The runtime for the downstream stance detection is about three GPU hours. The runtime for the remaining analyses is negligible.

\paragraph{Number of Parameters}
The total number of parameters in BERTje is 109,139,715.

\paragraph{Validation Performance}
The best, selected finetuned BERTje model has a loss of 0.41114.
%Corresponding validation performance for each reported test result (*)

\paragraph{Hyperparameter Search}
For the BERTje model, we use the same hyperparameters and search strategy as \cite{wijnholds-etal-2021-sicknl}, including: 

- num\_train\_epochs=20 

- per\_device\_train\_batch\_size=16

- per\_device\_eval\_batch\_size=64

- warmup\_steps=250     

- weight\_decay=0.01     

Among the 20 epochs, we select the one that has the lowest validation loss.

\paragraph{Cost of OpenAI's GPT-3.5 Turbo API}
1.13 US dollars.

\newpage
\section{The Survey Instruments and Hypotheses}
\label{app:chap7:survey}
The following survey items from LISS are used to construct the hypotheses in TER. English translation are in \textit{italics}:

1. Een werkende moeder kan niet zo'n warme en hechte relatie met haar kinderen hebben als een moeder die niet werkt. \textit{A working mother's relationship with her children cannot be just as close and warm as that of a non-working mother}

2. Een kind dat nog niet naar school gaat zal er waarschijnlijk onder lijden als zijn of haar moeder werkt. \textit{A child that is not yet attending school is likely to suffer the consequences if his or her mother has a job.}

3. Al met al lijdt het gezinsleven er onder als de vrouw een volledige baan heeft. \textit{Overall, family life suffers the consequences if the mother has a full-time job.}

4. Alleen de man moet bijdragen aan het gezinsinkomen. \textit{Only men should contribute to the family income.}

5. De man moet het geld verdienen, de vrouw moet voor het huishouden en het gezin zorgen. \textit{The father should earn money, while the mother takes care of the household and the family.}

6. Mannen zouden niet een groter deel van het huishoudelijk werk moeten doen dan nu het geval is. \textit{Fathers ought not to do more in terms of household work than they do at present.}

7. Mannen zouden niet meer moeten doen aan de verzorging van de kinderen dan nu het geval is. \textit{Fathers ought not to do more in terms of childcare than they do at present.}

8. Een vrouw is geschikter om kleine kinderen op te voeden dan een man. \textit{A woman is more suited to rearing young children than a man.}

9. Voor een meisje is het eigenlijk toch niet zo belangrijk als voor een jongen om een goede schoolopleiding te krijgen. \textit{It is actually less important for a girl than for a boy to get a good education.}

10. Jongens kun je in het algemeen vrijer opvoeden dan meisjes. \textit{Generally speaking, boys can be reared more liberally than girls.}

11. Het is onnatuurlijk als vrouwen in een bedrijf leiding uitoefenen over mannen. \textit{It is unnatural for women in firms to have control over men.}

Where survey items are not used, the following statement is adopted:

Ik ben voorstander van de traditionele rolverdeling tussen mannen en vrouwen. \textit{I am in favour of traditional gender role division.}

\section{Examples of Tweets of Different Stances towards Traditional Gender Role Division}
\label{app:chap7:examples}
See below for examples of Tweets implying different stances towards traditional gender role division. English translation in \textit{italics}.

\textbf{Against} traditional gender role division: ``Ik zal me altijd ten volle inzetten voor de rechtvaardige strijd voor het vervolmaken van de gelijkwaardige positie van de vrouw in onze samenleving. Dan reken ik ook op de steun van alle mannen.'' \textit{I will always be fully committed to the just struggle to perfect the equal position of women in our society. Then I also count on the support of all men.}

\textbf{In favour of} traditional gender role division: ``Zien we nog het voorrecht en het bijzondere van het moederschap? Ik zou bijna geen werk van een man kunnen bedenken wat zo hoog reikt als het moederschap.'' \textit{Do we still see the privilege and specialness of motherhood? I could hardly think of any work of a man that reaches as high as motherhood.}

\textbf{Neutral} towards or unrelated to traditional gender role division: ``Opnieuw het misverstand dat legalisering van abortus tot meer abortus leidt. Het tegendeel is het geval. Legalisering leidt niet tot meer abortus, maar wel tot meer veilige abortus en minder sterfte onder vrouwen.'' \textit{Again the misunderstanding that legalisation of abortion leads to more abortion. The opposite is the case. Legalisation does not lead to more abortion, but to more safe abortion and less death among women.}

\end{document}